\documentclass{article}
\usepackage{iclr2027_conference,times}
\usepackage{iftex}
\ifPDFTeX
  \usepackage[T1]{fontenc}
\else
  \usepackage{fontspec}
\fi
\usepackage{amsmath,amssymb,booktabs,array,tabularx,graphicx,xcolor,float}
\usepackage{multirow}
\usepackage[ruled,vlined,linesnumbered]{algorithm2e}
\usepackage{tikz}
\usetikzlibrary{arrows.meta}
\usepackage{wrapfig}
\usepackage{url,hyperref,etoolbox,needspace,capt-of,colortbl,placeins}
\hypersetup{unicode=true,colorlinks=true,citecolor=blue,linkcolor=blue,urlcolor=blue,
  pdftitle={MMPostTrainBench: Benchmarking Autonomous Research for Multimodal Post-Training},
  pdfauthor={Yuxin Liu, Yuxuan Wang, Zhenxin Lei, Lingchen Meng, Yuchong Sun, Junming Lin, Hongcheng Liu, Yunfei Chu, Qize Yang, Jin Xu, Lei Zhang, Zhendong Mao}}
\newcommand{\benchmark}{MMPostTrainBench}
\newcommand{\harness}{\mbox{MMResearch}}

\definecolor{milegain}{HTML}{286C65}
\definecolor{mileloss}{HTML}{925144}
\definecolor{milelight}{HTML}{F2F4F6}
\newcommand{\signedgain}[1]{\ifdim #1pt<0pt\textcolor{mileloss}{#1}\else\textcolor{milegain}{#1}\fi}
\newcommand{\scorecell}[2]{\shortstack{#1\\{\fontsize{8}{9}\selectfont\signedgain{#2}}}}
\newcommand{\miletablestyle}{\centering\small
  \setlength{\tabcolsep}{3.5pt}\renewcommand{\arraystretch}{1.12}
  \setlength{\abovecaptionskip}{4pt}\setlength{\belowcaptionskip}{5pt}}
\newcolumntype{Y}{>{\raggedright\arraybackslash}X}
\title{MMPostTrainBench: Benchmarking Autonomous Research for Multimodal Post-Training}
\author{%
  \begin{minipage}{\dimexpr\textwidth-2\tabcolsep\relax}
  \raggedright\normalfont\normalsize
  \mbox{\textbf{Yuxin Liu}\textsuperscript{1}}\quad \mbox{\textbf{Yuxuan Wang}\textsuperscript{2}}\quad \mbox{\textbf{Zhenxin Lei}\textsuperscript{3}}\quad \mbox{\textbf{Lingchen Meng}\textsuperscript{2}}\quad \mbox{\textbf{Yuchong Sun}\textsuperscript{2}} \\[2pt]
  \mbox{\textbf{Junming Lin}\textsuperscript{4}}\quad \mbox{\textbf{Hongcheng Liu}\textsuperscript{5}}\quad \mbox{\textbf{Yunfei Chu}\textsuperscript{2}}\quad \mbox{\textbf{Qize Yang}\textsuperscript{2}}\quad \mbox{\textbf{Jin Xu}\textsuperscript{2}} \\[2pt]
  \mbox{\textbf{Lei Zhang}\textsuperscript{1,*}}\quad \mbox{\textbf{Zhendong Mao}\textsuperscript{1}} \\[5pt]
  \textsuperscript{1}University of Science and Technology of China\quad\textsuperscript{2}Alibaba Token Hub, Alibaba Group\\[2pt]
  \textsuperscript{3}University of the Chinese Academy of Sciences\\[2pt]
  \textsuperscript{4}Tsinghua University\quad\textsuperscript{5}Shanghai Jiao Tong University\\[2pt]
  \textsuperscript{*}Corresponding author
  \end{minipage}%
}
\iclrfinalcopy
\makeatletter
\patchcmd{\@maketitle}
  {Published as a conference paper at ICLR 2027}
  {Preprint}
  {}
  {\PackageError{mmptb}{Preprint header patch failed}{Check iclr2027_conference.sty.}}
\makeatother
\begin{document}
\raggedbottom
\maketitle
\vspace{-8pt}
\begin{abstract}
Autonomous research seeks sustained model improvements through iterative experimentation and feedback. LLM agents show promise in automating machine learning and language-model post-training, but their ability to sustain multimodal improvement remains unclear. We introduce \benchmark{}, a benchmark spanning eight tasks in image, audio, video, and joint audio-video understanding and image-grounded software repair. Agents operate from a common base model within fixed budgets, using development feedback before independent evaluation of their submitted models. Evaluation covers target and non-target model outcomes, iterative model improvement and selection, and research integrity. Across all eight tasks, 52.1\% of model--task means fall below the base, and evaluated submissions also exhibit non-target regressions. Model performance does not consistently improve across research iterations, and agents do not reliably select the best evaluated candidate for submission; final submissions trail that candidate by up to 5.38 percentage points. Extending autonomous research from text-only to multimodal tasks introduces additional sources of error in perception, cross-modal alignment, and temporal grounding. The observed regressions and selection gaps highlight the need to balance targeted improvements with non-target capability preservation and to retain gains across research iterations. These requirements motivate \harness{}, a multimodal research framework that connects media-grounded evidence to hypotheses and interventions, carries findings across rounds through hierarchical memory, and retains candidates using development evaluation. Added to existing code-agent runtimes, it improves submitted-model accuracy by up to 7.75 percentage points for Claude Opus 4.8 with Claude Code and 2.33 points for GPT-5.6-sol with Codex.
\end{abstract}
\begingroup
\setlength{\intextsep}{5pt}
\begin{figure}[H]
\centering
\includegraphics[width=0.88\linewidth]{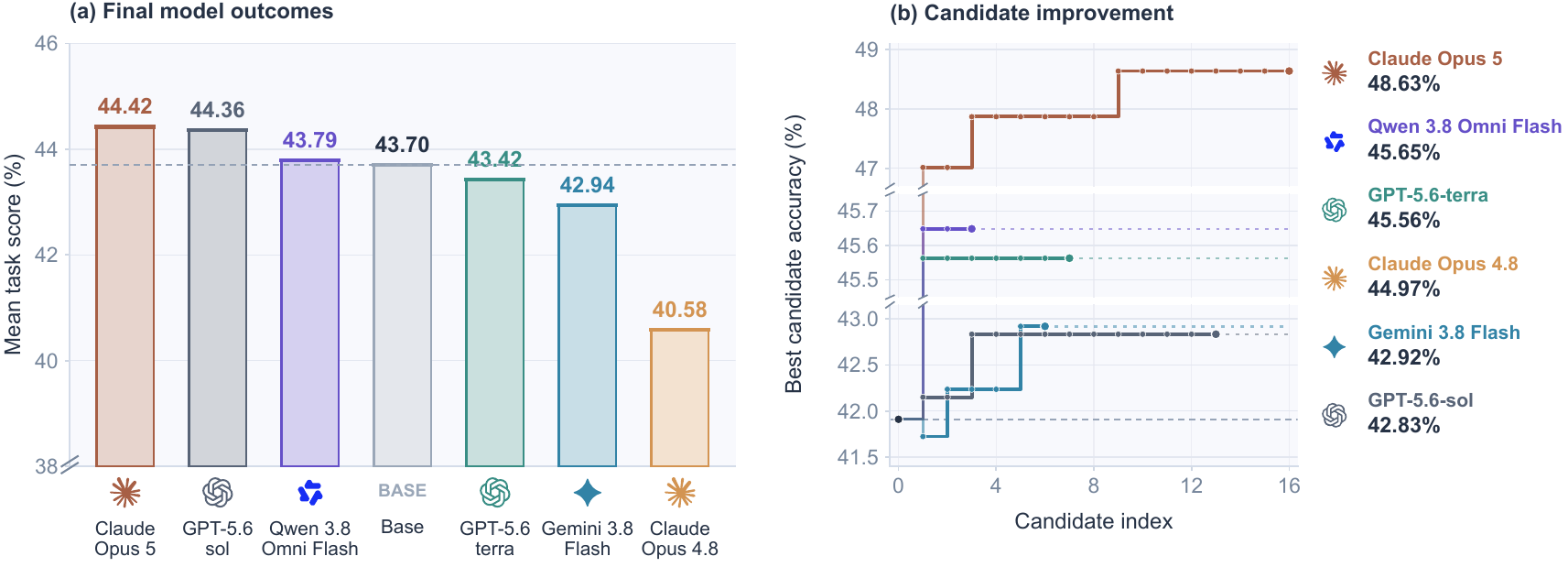}
\caption{\textbf{MMPostTrainBench at a glance.} (a) \textbf{Final model outcomes:} post-training effectiveness measured by mean task score across all eight tasks relative to the common base model (Table~\ref{tab:main}). (b) \textbf{Iterative model improvement:} best-so-far held-out candidate accuracy on MMMU-Pro; labels report the best candidate score.}
\label{fig:key-results}
\end{figure}

\endgroup
\Needspace{7\baselineskip}
\begingroup
\patchcmd{\section}{1.5ex plus 0.3ex minus0.2ex}{1.5ex}{}
  {\PackageError{mile}{Introduction heading spacing patch failed}{Check the current section definition.}}
\section{Introduction}
\label{sec:intro}
A central challenge in autonomous machine-learning research is to translate experimental feedback into effective model improvements and accumulate useful knowledge across successive experiments. Research agents tackle this challenge with code, tools, and feedback for experiments, language-model post-training, and training-data research~\citep{mlagentbench,mlebench,rebench,aiscientist,posttrainbench,rsibenchdata}. Beyond text-only post-training, agents must connect evidence in source media with language and other modalities. Clutter, background sounds, and incomplete descriptions in media collected in the wild make relevant evidence difficult to isolate. This raises a broader question: \emph{Can research agents autonomously improve how models see, hear, and reason about multimodal inputs---and sustain those improvements as experimentation continues?}

Autonomous multimodal post-training couples three challenges. \textbf{(1) Multimodal Error Diagnosis.} In visual reasoning, chain-of-thought prompting can degrade performance on perception-heavy tasks~\citep{mmecot}. Agents must inspect source media to distinguish missed or misinterpreted evidence from faulty inference before choosing an intervention. \textbf{(2) Data Curation and Modality Alignment.} Selecting visual instructions for vulnerable sample groups can improve robustness~\citep{ards}. Agents must also preserve semantic and temporal alignment: labels should match observable content, and paired audio-video segments should retain their timing. \textbf{(3) Budgeted Multimodal Experimentation.} Autonomous language-model post-training evaluations report early termination without valid final checkpoints~\citep{posttrainbench}. Multimodal runs must also budget for frame sampling and audio processing alongside training, candidate evaluation, and final delivery. Effective research must therefore connect media evidence and alignment decisions to the improvements ultimately delivered.

We introduce \benchmark{} (Multimodal Post-Training Benchmark)\footnote{Benchmark code and evaluation setup: \url{https://github.com/sod1010/MMPostTrainBench}.}, which evaluates autonomous research on eight tasks spanning image, audio, video, and joint audio-visual understanding, as well as multimodal code tasks (Table~\ref{tab:tasks}; Figure~\ref{fig:protocol}). Task queries specify a target capability and benchmark, such as audio reasoning on MMAR, and require training pipelines to process the corresponding media. Under a common resource budget, agents select data sources, modality mixtures, supervision, and training strategies in response to development feedback. Harbor-compatible task packages provide this feedback during research and reserve separate held-out tests for final evaluation. Target and non-target evaluations assess whether a targeted update preserves other capabilities.

We assess \textbf{model outcomes}, \textbf{iterative model improvement}, and \textbf{research integrity}. Across six research-agent configurations on eight tasks, 52.1\% of model--task means fall below the base, and submissions also exhibit non-target regressions. Figure~\ref{fig:key-results} summarizes final outcomes and best-so-far candidate improvement. Final submissions trail the best evaluated candidates by up to 5.38 percentage points across recorded candidate pools. Integrity checks flag suspected misuse of evaluation data during training. Producing a strong candidate, selecting it reliably, and retaining other multimodal capabilities thus remain distinct requirements.

These findings motivate our proposed multimodal research framework, \harness{}. Building on persistent research orchestration~\citep{arbor}, it augments existing code-agent runtimes with \textbf{multimodal evidence reasoning}, \textbf{hierarchical memory}, and \textbf{evaluation-guided model selection}. Observations are anchored to source media and linked to hypotheses, interventions, and development outcomes; memory carries successful recipes and failure records across rounds; model selection retains a candidate for final delivery. These mechanisms connect diagnosis, accumulated knowledge, and submission decisions within a seven-step research loop. In the evaluated configurations, adding \harness{} improves submitted-model accuracy by up to 7.75 percentage points for Claude Opus 4.8 with Claude Code and 2.33 points for GPT-5.6-sol with Codex.

Our contributions are threefold: \textbf{(1)} We introduce \benchmark{}, which evaluates autonomous post-training across eight multimodal tasks through model outcomes, iterative model improvement, and research integrity. \textbf{(2)} We characterize non-target regressions, candidate degradation, and submission-selection failures in current research agents. \textbf{(3)} We propose \harness{}, which combines multimodal evidence reasoning, hierarchical memory, and evaluation-guided model selection to improve submitted models within existing code-agent runtimes.

\par
\endgroup
\section{Related Work}
\label{sec:related}

\paragraph{Autonomous Research Benchmarks.}
Autonomous research benchmarks evaluate progress at different levels of an AI system. Task-level evaluations measure agents' ability to conduct ML experiments and solve AI R\&D problems~\citep{mlagentbench,mlebench,rebench}. Other evaluations focus on the supporting harness and infrastructure~\citep{harnessdev,phibench}, or on the agent's ability to learn from broad goals and interaction feedback~\citep{aspire,s3gym}. For autonomous post-training, the object of improvement is the target model: PostTrainBench permits end-to-end language-model post-training, whereas RSIBench-Data fixes the learning stack to isolate training-data research~\citep{posttrainbench,rsibenchdata}. \benchmark{} extends this setting to multimodal tasks, assessing final target and non-target performance, iterative candidate quality and submission choice, and research integrity.

\paragraph{Research Agent Harnesses.}
Research agent harnesses must organize experiments and make accumulated evidence useful for subsequent decisions. End-to-end workflows coordinate idea generation, implementation, and training~\citep{aiscientist,aibuildai}, while search-based systems structure the choice of experiments~\citep{aide}. As experimentation continues, preserving and reusing research state becomes equally important. Hierarchical knowledge retrieval and persistent hypothesis trees connect earlier findings to later decisions~\citep{aibuildai2,arbor}; discovery replay and reusable environment memory serve a related role in autonomous exploration~\citep{dreamrsi,rsiagent}. ModularRSI refactors Terminus-2 into behavioral modules and evolves their source functions~\citep{wu2026modularrsi}. Building on research orchestration and knowledge reuse, \harness{} focuses on improving multimodal models through post-training, connecting multimodal evidence-based diagnosis with training interventions and model selection.

\Needspace{8\baselineskip}
\section{Method}
\label{sec:benchmark}
\begin{figure}[t]
\centering
\includegraphics[width=1.0\linewidth]{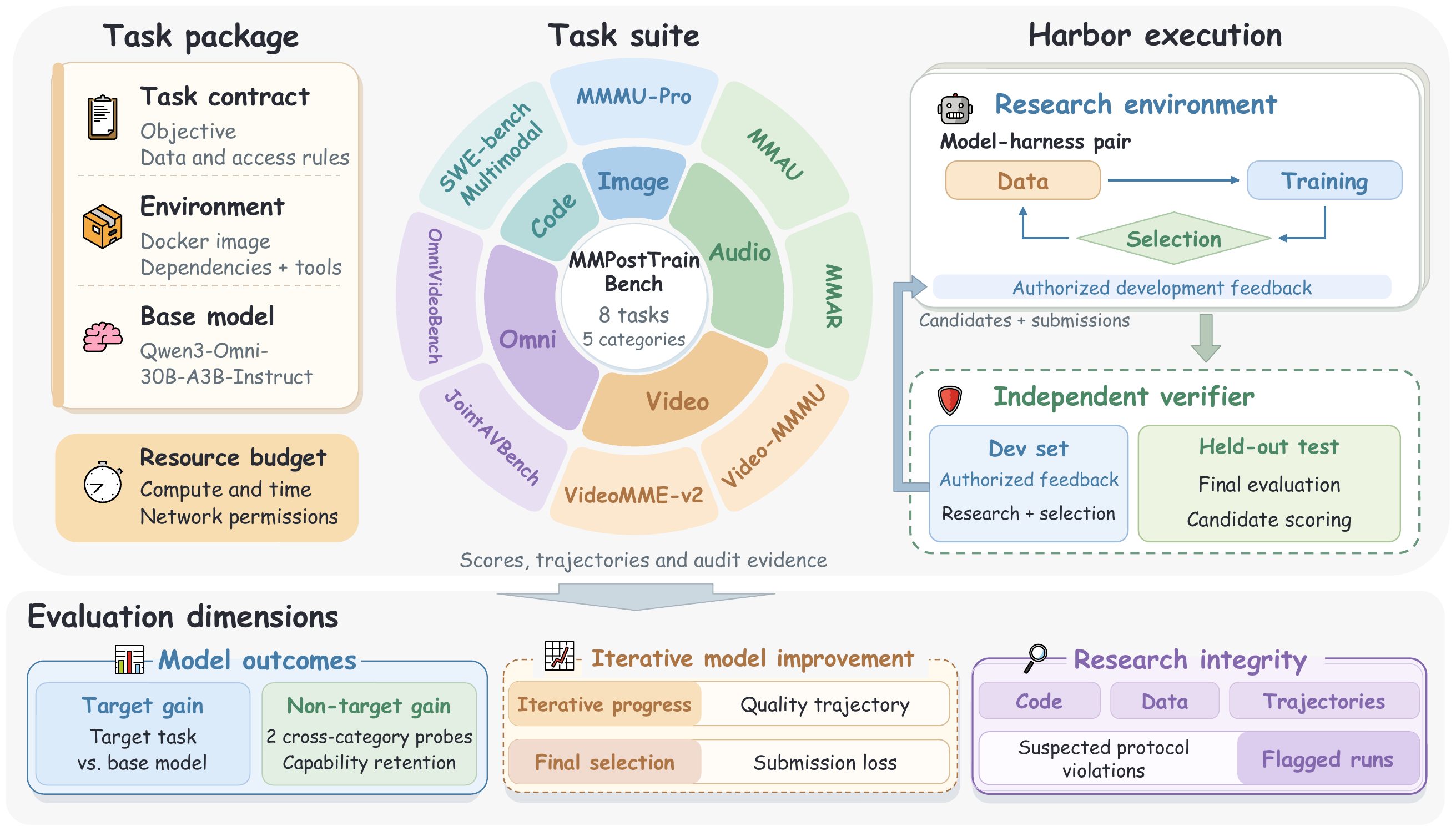}
\caption{\textbf{MMPostTrainBench benchmark overview.} Agents conduct post-training from a common base model on eight multimodal tasks within fixed budgets. To limit evaluation leakage, execution and verification run in separate containers, research feedback is restricted to authorized development data, and final tests remain sealed. Submitted models, candidate trajectories, and audits support evaluation of model outcomes, iterative model improvement, and research integrity.}
\label{fig:protocol}
\end{figure}

\benchmark{} evaluates autonomous multimodal post-training through a controlled research environment and three complementary evaluation dimensions (Figure~\ref{fig:protocol}). We describe the task setting and evaluation protocol before introducing our proposed research framework, \harness{}.

\subsection{Task Definition and Research Process}
\label{sec:task-definition}
\paragraph{Research task.}
A task assigns the agent a capability to improve, a starting model, and a bounded research environment. Following the artifact-oriented setup of autonomous post-training and research benchmarks~\citep{posttrainbench,rsibenchdata,arbor}, we represent task $b$ as
\begin{equation}
\mathcal{T}_b=(q_b,\theta_0,\mathcal{D}^{\mathrm{adm}}_b,F_b^{\mathrm{dev}},E_b^{\mathrm{test}},\mathcal{B}_b).
\label{eq:research-task}
\end{equation}
Here $q_b$ specifies the target capability and success criterion; $\theta_0$ is the common base model; $\mathcal{D}^{\mathrm{adm}}_b$ specifies permitted training-data sources and uses; $F_b^{\mathrm{dev}}$ returns authorized development feedback; $E_b^{\mathrm{test}}$ independently evaluates frozen models; and $\mathcal{B}_b$ specifies compute allocation and the research time limit. Research configuration $a$ consists of a frontier model and its execution runtime.

\paragraph{Autonomous research.}
Within loop $r$, the agent chooses interventions from its accumulated history rather than following a prescribed training recipe. Suppressing $a,b,r$ for clarity, candidate production and feedback can be written as
\begin{align}
u_t&=\pi_a(q_b,\mathcal{H}_{<t}), &
\theta_t&=\operatorname{Execute}(u_t;\theta_0,\mathcal{C}_{<t}),\nonumber\\
f_t&=F_b^{\mathrm{dev}}(\theta_t), &
\mathcal{H}_{\le t}&=\mathcal{H}_{<t}\cup\{(u_t,\theta_t,f_t,\ell_t)\}.
\label{eq:research-process}
\end{align}
An intervention $u_t$ specifies data construction, a training recipe, or a model-side operation using permitted prior candidates $\mathcal{C}_{<t}$; $\ell_t$ records execution diagnostics and provenance. The initial history contains permitted task materials, and later observations include only information available during research. The agent decides how many experiments to attempt and which completed model $S_{a,b,r}=\theta^{\mathrm{sub}}_{a,b,r}$ to submit before the budget expires.

\begin{table}[!htbp]
\centering\renewcommand{\arraystretch}{1.15}\small
\setlength{\tabcolsep}{4pt}
\caption{Eight post-training tasks, their domains, and target model abilities.}
\label{tab:tasks}
\begin{tabular}{@{}lll@{}}
\toprule
Task & Domain & Model Ability \\
\midrule
MMMU-Pro~\citep{mmmupro} & Image & Image-grounded multidisciplinary reasoning \\
\addlinespace[3pt]
MMAU~\citep{mmau} & Audio & Audio understanding and reasoning \\
MMAR~\citep{mmar} & Audio & Audio reasoning \\
\addlinespace[3pt]
Video-MMMU~\citep{videommmu} & Video & Video-grounded multidisciplinary reasoning \\
VideoMME-v2~\citep{videommev2} & Video & Video understanding \\
\addlinespace[3pt]
JointAVBench~\citep{jointavbench} & Omni & Joint audio-video understanding \\
OmniVideoBench~\citep{omnivideobench} & Omni & Joint audio-video understanding and reasoning \\
\addlinespace[3pt]
SWE-bench Multimodal~\citep{swebench_multimodal} & Code & Image-grounded software repair \\
\bottomrule
\end{tabular}
\end{table}

\subsection{Research Environment and Protocol}
\label{sec:environment}
\paragraph{Task environment.}
Each Harbor-compatible task package specifies the task objective, designated base model, data-use rules, tool interfaces, and resource limits. The research environment provides a writable workspace and the dependencies required for model training and evaluation. Within the task's permissions, agents can inspect multimodal media, construct training examples, modify training code and configurations, execute experiments, and retain candidate models. Harbor's separate-verifier interface and the Docker driver separate research execution from independent verification.

\paragraph{Development feedback.}
Agents query $F_b^{\mathrm{dev}}$ to obtain candidate scores, permitted evaluation examples, and execution diagnostics. These observations support error diagnosis and subsequent decisions about data construction, training configurations, and candidate selection. The task's data-use rules determine which materials may serve as training supervision; access to evaluation examples for diagnosis does not itself authorize their use for training. Development feedback excludes final held-out test instances, reference answers, and scores, which remain withheld during research.

\paragraph{Submission and verification.}
At the end of research, the selected model's weights, configuration, and loading dependencies are frozen and transferred to the independent verifier. The final evaluator $E_b^{\mathrm{test}}$ assesses the base and submitted models using the same fixed held-out test instances, preprocessing, inference interface, and scoring procedure. This matched evaluation provides the basis for measuring changes in model performance. Data provenance, tool traces, and candidate records connect the submitted model to its research process and support the integrity checks defined in Section~\ref{sec:evaluation-dimensions}.

\subsection{Evaluation Dimensions}
\label{sec:evaluation-dimensions}
We evaluate three complementary dimensions: model outcomes, iterative model improvement, and research integrity. Let $S=\theta^{\mathrm{sub}}_{a,b}$ denote the submitted model and $H_b(\theta)\in[0,1]$ its raw task score: answer accuracy for understanding tasks and resolved-instance rate for SWE-bench Multimodal.

\noindent\textbf{Model outcomes.}
We compare submissions with the base on the target task and two prespecified non-target probes from other categories. Target change is $\Delta_{a,b}=100[\widehat{s}_{a,b}-H_b(\theta_0)]$ percentage points, where $\widehat{s}_{a,b}$ is the mean of three loop scores, with flagged loops reset to the base before averaging. Non-target changes assess transfer and capability retention. Aggregate comparisons weight all eight tasks equally. For SWE-bench Multimodal, Table~\ref{tab:main} reports submitted-model means relative to a base resolved-instance rate of 2.29\%.

\noindent\textbf{Iterative model improvement.}
We assess iterative model improvement from two perspectives. \textbf{(1) Iterative progress:} whether further experiments improve candidate quality or lead to stagnation, degradation, or fluctuations, as reflected by candidate trajectories evaluated retrospectively on the same held-out test set. \textbf{(2) Final selection:} whether the final submission retains the best performance among the evaluated candidates produced before submission. Let $\mathcal{C}^{\mathrm{rec}}_{a,b}$ denote this candidate set. Submission loss is defined as
\begin{equation}
R^{\mathrm{rec}}_{a,b}=100\!\left[\max_{\theta\in\mathcal{C}^{\mathrm{rec}}_{a,b}\cup\{S\}}H_b(\theta)-H_b(S)\right].
\label{eq:submission-loss}
\end{equation}
A positive value indicates that the final submission underperforms the best evaluated candidate, with the gap measured in percentage points.

\noindent\textbf{Research integrity.}
Benchmark scores may be inflated by contamination or misuse of evaluation information. Audit categories cover training-data contamination, unauthorized workspace or data access, prohibited split use, and misuse of evaluation information. Retained code, data provenance, and research trajectories provide evidence for these checks. Results are summarized by the number of runs flagged for suspected violations.

\subsection{MMResearch: Multimodal Research Framework}
\label{sec:harness}
\begin{figure}[t]
\centering
\includegraphics[width=1.0\linewidth]{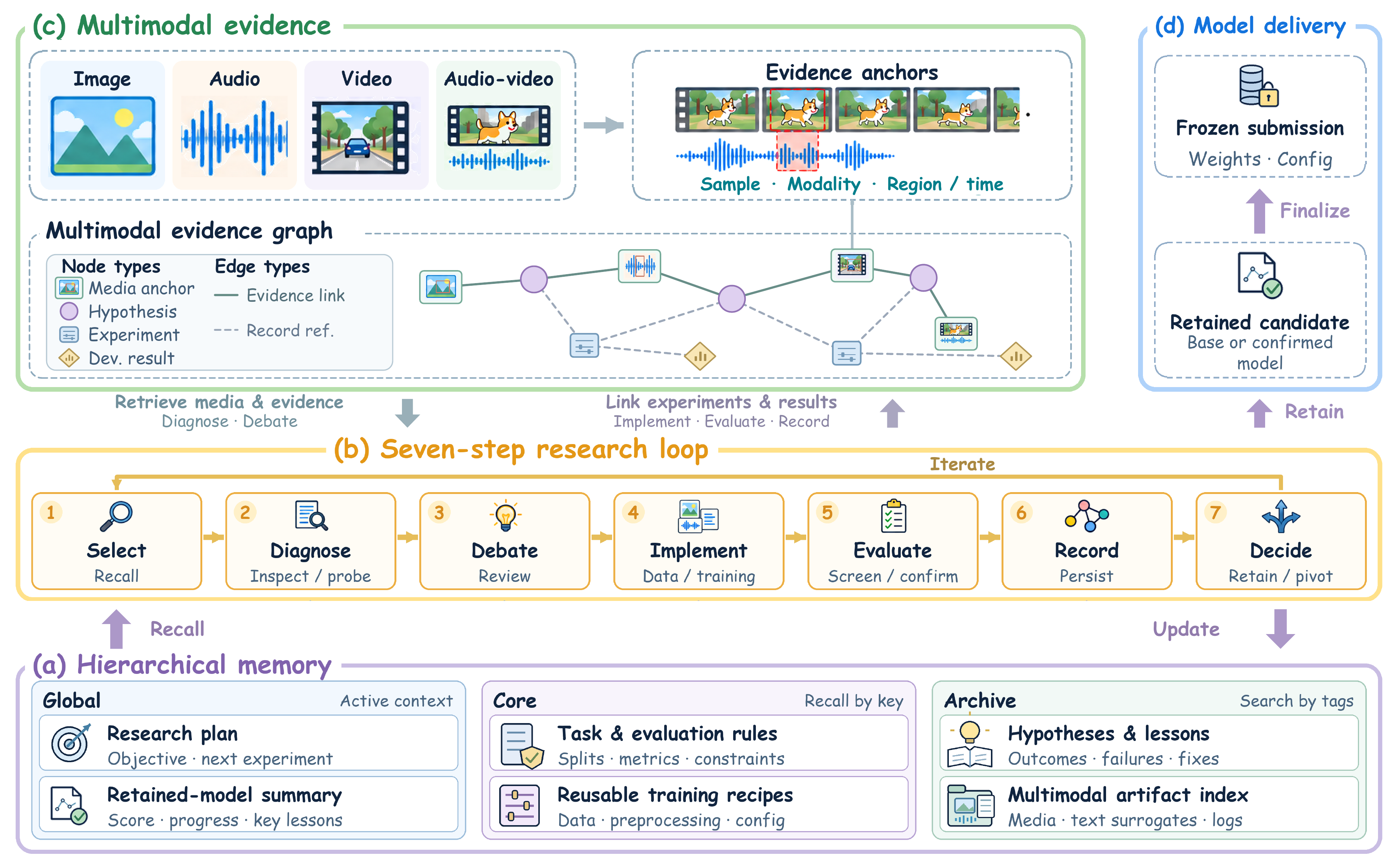}
\caption{\textbf{MMResearch framework.} MMResearch coordinates a seven-step research loop (b), using hierarchical memory (a) to guide experimentation and multimodal evidence (c) to connect observations with experimental decisions, then freezes the retained candidate for delivery (d).}
\label{fig:harness}
\end{figure}

\harness{} adds a research controller to existing code-agent runtimes, retaining their native tools and action loops while coordinating experiments, memory, and model delivery. Figure~\ref{fig:harness} shows how hierarchical memory (a) and multimodal evidence (c) support the seven-step research loop (b), which concludes with model delivery (d). The evidence-anchor and graph design in (c) specifies how localized media observations connect to research hypotheses and experimental outcomes.

\noindent\textbf{Seven-Step Research Loop.}
Each round turns a research direction into an evaluated candidate and a decision (Figure~\ref{fig:harness}b). \textsc{Select} chooses a data- or model-side direction from prior hypotheses and failure records. \textsc{Diagnose} inspects authorized multimodal inputs to investigate the suspected failure. \textsc{Debate} reviews the diagnosis, intervention, feasibility, and risks. \textsc{Implement} executes the intervention and saves a candidate. \textsc{Evaluate} screens and confirms candidates on development data. \textsc{Record} stores configurations, results, and rationales in hierarchical memory. \textsc{Decide} uses confirmation results to promote the candidate, refine the approach, or pivot to another direction; a proxy improvement alone does not justify promotion. A separate checkpoint tracks the retained model, initially the base. Before the budget expires, the controller checks required records and freezes the retained model's weights and configuration for delivery.

\noindent\textbf{Multimodal Evidence.}
The agent inspects authorized image, audio, video, and audio-video inputs through native perception or configured tools. In the \emph{multimodal evidence graph} design (Figure~\ref{fig:harness}c), each \emph{evidence anchor} identifies a source sample, its modality, and a relevant region or time span. When finer localization is unavailable, the anchor refers to the whole image or clip. Direct observations are stored separately from the hypotheses inferred from them. Links connect observations, hypotheses, training interventions, and development-evaluation outcomes, making diagnoses traceable to the underlying media. \textsc{Diagnose} and \textsc{Debate} use the linked observations to review explanations; \textsc{Record} preserves the evidence and outcome links for subsequent rounds. Later rounds can follow these links back to the source media and prior interventions to reassess an earlier diagnosis in light of development results. Held-out test results remain outside this research feedback loop.

\noindent\textbf{Hierarchical Memory.}
\harness{} preserves research context across rounds in three memory tiers (Figure~\ref{fig:harness}a). \emph{Global} maintains the active research plan and a summary of the retained model. \emph{Core} stores task and evaluation rules and reusable training recipes. \emph{Archive} indexes hypotheses, outcomes, failures, and media artifacts. \textsc{Select} retrieves relevant records at the start of a round, and \textsc{Record} updates them with its results and lessons, allowing later experiments to build on earlier findings.

\section{Experiments}
\label{sec:results}
\subsection{Experimental Setup}
\noindent\textbf{Tasks and Resources.}\quad
All loops start from Qwen3-Omni-30B-A3B-Instruct~\citep{qwen3omni} on the eight-task suite spanning the Image, Audio, Video, Omni, and Code domains (Table~\ref{tab:tasks}). Training uses eight GPUs, with a default limit of 24 hours per loop.

\noindent\textbf{Research Agents.}\quad
We evaluate six research-model/runtime configurations: GPT-5.6-sol and GPT-5.6-terra with Codex; Claude Opus 5 and Claude Opus 4.8 with Claude Code; Gemini 3.8 Flash with Gemini CLI; and Qwen 3.8 Omni Flash with Qwen Code. These models direct experiments through their native runtimes, while Qwen3-Omni is the shared model being post-trained.

\noindent\textbf{Evaluation Protocol.}\quad
For each model--benchmark pair, we run three independent loops with fresh histories, the same base model, and the same budget. Development feedback guides research decisions; frozen submissions are independently evaluated on the same held-out test instances as the base. Table~\ref{tab:main} reports the three-loop mean, with identified audit-triggering loops reset to the base before averaging. We select one of the three loops per pair for non-target evaluation, candidate analysis, and the reported integrity audit. Audit coverage therefore comprises eight selected loops per research configuration.

\Needspace{8\baselineskip}
\subsection{Main Results}
\paragraph{Model outcomes.}
Across all eight tasks, 52.1\% of the 48 model--task means fall below the base; the largest mean gain is 0.73 pp (Table~\ref{tab:main}). Qwen 3.8 Omni Flash gains 2.80 pp on MMMU-Pro but loses 2.26 pp on OmniVideoBench. Gemini 3.8 Flash leads on MMAR yet regresses on both video tasks. Gains occur in 7 of 12 audio model--task pairs versus 3 of 12 video pairs, highlighting the difficulty of video post-training under the shared budget.

\begin{table}[!htbp]
\miletablestyle
\setlength{\tabcolsep}{1.4pt}
\caption{\textbf{Final model outcomes on MMPostTrainBench.} Three-loop mean task score (\%) and change from base (pp). Harnesses appear below model names; bold marks the best result per row. The aggregate weights all eight tasks equally.}
\label{tab:main}

\begin{tabular*}{\linewidth}{@{}>{\fontsize{8}{9}\selectfont}l@{\hspace{8pt}}>{\fontsize{8}{9}\selectfont}l@{\hspace{4.5pt}}c@{\extracolsep{\fill}}cccccc@{}}
\toprule
\multirow{2}{*}[-0.65ex]{\textbf{Category}} & \multirow{2}{*}[-0.65ex]{\textbf{Benchmark}} & \multirow{2}{*}[-0.65ex]{\shortstack{\textbf{Base}\\\textbf{model}}} &
\shortstack{\textbf{Claude}\\\textbf{Opus 5}} & \shortstack{\textbf{GPT-5.6}\\\textbf{sol}} &
\shortstack{\textbf{Qwen 3.8}\\\textbf{Omni Flash}} & \shortstack{\textbf{GPT-5.6}\\\textbf{terra}} &
\shortstack{\textbf{Gemini 3.8}\\\textbf{Flash}} & \shortstack{\textbf{Claude}\\\textbf{Opus 4.8}} \\
\cmidrule(lr){4-9}
 & & &
{\fontsize{8}{9}\selectfont Claude Code} &
{\fontsize{8}{9}\selectfont Codex} &
{\fontsize{8}{9}\selectfont Qwen Code} &
{\fontsize{8}{9}\selectfont Codex} &
{\fontsize{8}{9}\selectfont Gemini CLI} &
{\fontsize{8}{9}\selectfont Claude Code} \\
\midrule
\textit{Image} & MMMU-Pro & 41.91 & \scorecell{43.26}{+1.35} & \scorecell{41.88}{-0.03} & \textbf{\scorecell{44.71}{+2.80}} & \scorecell{43.00}{+1.09} & \scorecell{41.88}{-0.03} & \scorecell{41.72}{-0.19} \\
\arrayrulecolor{black!18}\specialrule{0.25pt}{2pt}{2pt}\arrayrulecolor{black}
\multirow[c]{2}{*}[-4.85pt]{\textit{Audio}} & MMAU & 48.60 & \scorecell{50.25}{+1.65} & \textbf{\scorecell{50.30}{+1.70}} & \scorecell{50.00}{+1.40} & \scorecell{49.40}{+0.80} & \scorecell{50.00}{+1.40} & \scorecell{44.01}{-4.59} \\
 & MMAR & 72.40 & \scorecell{72.16}{-0.24} & \scorecell{72.31}{-0.09} & \scorecell{72.01}{-0.39} & \scorecell{73.00}{+0.60} & \textbf{\scorecell{73.50}{+1.10}} & \scorecell{66.77}{-5.63} \\
\arrayrulecolor{black!18}\specialrule{0.25pt}{2pt}{2pt}\arrayrulecolor{black}
\multirow[c]{2}{*}[-4.85pt]{\textit{Video}} & Video-MMMU & 60.00 & \textbf{\scorecell{60.89}{+0.89}} & \scorecell{60.40}{+0.40} & \scorecell{53.96}{-6.04} & \scorecell{58.42}{-1.58} & \scorecell{49.01}{-10.99} & \scorecell{56.44}{-3.56} \\
 & VideoMME-v2 & 24.81 & \scorecell{23.40}{-1.41} & \scorecell{24.35}{-0.46} & \textbf{\scorecell{25.81}{+1.00}} & \scorecell{22.49}{-2.32} & \scorecell{23.21}{-1.60} & \scorecell{22.89}{-1.92} \\
\arrayrulecolor{black!18}\specialrule{0.25pt}{2pt}{2pt}\arrayrulecolor{black}
\multirow[c]{2}{*}[-4.85pt]{\textit{Omni}} & JointAVBench & 59.94 & \scorecell{59.91}{-0.03} & \scorecell{59.96}{+0.02} & \scorecell{63.90}{+3.96} & \scorecell{59.97}{+0.03} & \textbf{\scorecell{64.91}{+4.97}} & \scorecell{58.84}{-1.10} \\
 & OmniVideoBench & 39.61 & \scorecell{42.40}{+2.79} & \textbf{\scorecell{42.99}{+3.38}} & \scorecell{37.35}{-2.26} & \scorecell{39.02}{-0.59} & \scorecell{38.57}{-1.04} & \scorecell{31.86}{-7.75} \\
\arrayrulecolor{black!18}\specialrule{0.25pt}{2pt}{2pt}\arrayrulecolor{black}
\textit{Code} & \shortstack[l]{SWE-bench\\Multimodal} & 2.29 & \textbf{\scorecell{3.13}{+0.83}} & \scorecell{2.71}{+0.42} & \scorecell{2.57}{+0.28} & \scorecell{2.08}{-0.21} & \scorecell{2.43}{+0.14} & \scorecell{2.08}{-0.21} \\
\midrule
\multicolumn{2}{@{}l}{Mean task score (\%)} & 43.70 & \textbf{\scorecell{44.42}{+0.73}} & \scorecell{44.36}{+0.67} & \scorecell{43.79}{+0.09} & \scorecell{43.42}{-0.27} & \scorecell{42.94}{-0.76} & \scorecell{40.58}{-3.12} \\
\bottomrule
\end{tabular*}
\vspace{6pt}
\end{table}

\paragraph{Cross-modal capability retention.}
\label{sec:nontarget}
Non-target evaluation reveals capability loss beyond the optimization target (Table~\ref{tab:nontarget-results}). For example, Qwen's MMMU-Pro submission loses 9.88 pp on MMAR. All six SWE-bench Multimodal configurations improve MMMU-Pro scores by 0.12--1.67 pp while reducing MMAR scores by 0.20--1.00 pp (Appendix~\ref{app:nontarget-matrix}). These results show that image-grounded software-repair research can also affect audio understanding. Target accuracy and non-target retention therefore capture distinct aspects of post-training quality.

\begin{table}[!htbp]
\miletablestyle
\caption{\textbf{Cross-modal capability retention.} Cells show mean accuracy changes over the two indicated non-target probes (pp): Image/MMMU-Pro, Audio/MMAR, and Omni/JointAVBench. Underlining indicates the best result in each row.}
\label{tab:nontarget-results}
\fontsize{8}{9.2}\selectfont
\setlength{\tabcolsep}{2.5pt}
\renewcommand{\arraystretch}{1.20}
\begin{tabularx}{\linewidth}{@{}l@{\hspace{8pt}}l*{6}{>{\centering\arraybackslash}X}@{}}
\toprule
\textbf{\shortstack[l]{Target\\benchmark}} & \textbf{\shortstack[l]{Non-target\\modalities}} & \textbf{\shortstack{Claude\\Opus 5}} & \textbf{\shortstack{GPT-5.6\\sol}} & \textbf{\shortstack{GPT-5.6\\terra}} & \textbf{\shortstack{Gemini 3.8\\Flash}} & \textbf{\shortstack{Qwen 3.8\\Omni Flash}} & \textbf{\shortstack{Claude\\Opus 4.8}} \\
\midrule
MMMU-Pro & Audio, Omni & \underline{\signedgain{+4.08}} & \signedgain{+0.84} & \signedgain{-0.51} & \signedgain{+2.78} & \signedgain{-4.40} & \signedgain{-8.45} \\
MMAR & Image, Omni & \signedgain{-0.98} & \signedgain{-0.79} & \signedgain{-0.57} & \signedgain{-0.15} & \signedgain{+0.43} & \underline{\signedgain{+1.35}} \\
MMAU & Image, Omni & \signedgain{+1.71} & \signedgain{-1.06} & \signedgain{+1.26} & \signedgain{-1.43} & \signedgain{+0.04} & \underline{\signedgain{+4.75}} \\
Video-MMMU & Image, Audio & \signedgain{+1.18} & \signedgain{+1.01} & \signedgain{+0.39} & \underline{\signedgain{+1.86}} & \signedgain{-1.17} & \signedgain{-2.95} \\
VideoMME-v2 & Image, Audio & \underline{\signedgain{+0.82}} & \signedgain{+0.76} & \signedgain{-0.42} & \signedgain{-1.08} & \signedgain{-1.50} & \signedgain{-0.08} \\
JointAVBench & Image, Audio & \signedgain{-0.85} & \underline{\signedgain{+0.90}} & \signedgain{+0.34} & \signedgain{-15.31} & \signedgain{-3.37} & \signedgain{-5.24} \\
OmniVideoBench & Image, Audio & \signedgain{-0.28} & \signedgain{+0.50} & \underline{\signedgain{+0.84}} & \signedgain{+0.00} & \signedgain{-0.08} & \signedgain{-11.68} \\
\shortstack[l]{SWE-bench\\Multimodal} & Image, Audio & \signedgain{+0.34} & \signedgain{-0.29} & \signedgain{-0.24} & \signedgain{+0.24} & \underline{\signedgain{+0.50}} & \signedgain{-0.31} \\
\midrule
\rowcolor{milelight}
\multicolumn{2}{@{}l}{\textbf{Overall}} & \underline{\signedgain{+0.75}} & \signedgain{+0.23} & \signedgain{+0.14} & \signedgain{-1.64} & \signedgain{-1.20} & \signedgain{-2.83} \\
\bottomrule
\end{tabularx}
\end{table}

\FloatBarrier

\par\vspace*{6pt}
\noindent\begin{minipage}{\linewidth}
\begin{minipage}[t]{0.54\linewidth}
\vspace{0pt}
\noindent\textbf{Resource Usage.}\quad
Figure~\ref{fig:resource-usage} reports mean loop duration and agent token usage for the existing run set. GPT-5.6-terra averages 23.7 hours under the 24-hour budget, yet its mean task score remains 0.27 pp below the base (Table~\ref{tab:main}). Claude Opus 5 and GPT-5.6-sol achieve similar eight-task mean scores (44.42\% and 44.36\%), while Opus 5 uses about 31\% fewer tokens. Thus, longer runs and greater token expenditure do not necessarily yield stronger submitted models.
\end{minipage}\hfill
\begin{minipage}[t]{0.44\linewidth}
\vspace{0pt}
\centering
\includegraphics[width=\linewidth,trim=0 10bp 0 0,clip]{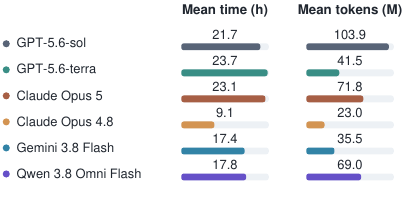}
\setlength{\abovecaptionskip}{2pt}
\setlength{\belowcaptionskip}{0pt}
\captionof{figure}{Mean time and tokens.}
\label{fig:resource-usage}
\end{minipage}
\end{minipage}\par

\begin{figure}[!htbp]
\centering
\begin{minipage}[t]{0.485\linewidth}
\vspace{0pt}
\centering
\includegraphics[width=0.94\linewidth]{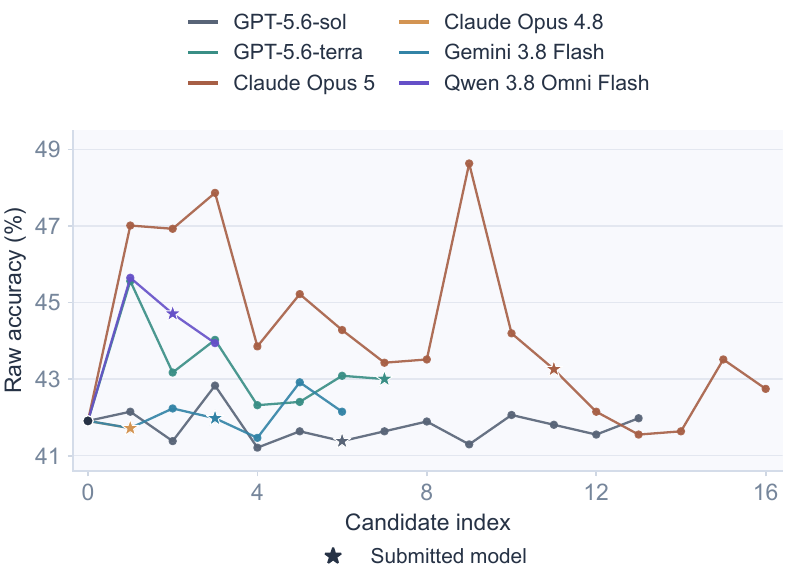}
\setlength{\abovecaptionskip}{3pt}
\setlength{\belowcaptionskip}{0pt}
\caption{\textbf{MMMU-Pro candidate accuracy.}}
\label{fig:selfevolve-results}
\end{minipage}\hfill
\begin{minipage}[t]{0.485\linewidth}
\vspace{0pt}
\centering
\setlength{\abovecaptionskip}{0pt}
\setlength{\belowcaptionskip}{4pt}
\captionof{table}{\textbf{Submission loss.} $R^{\mathrm{rec}}$ is best-prior minus submitted accuracy (pp).}
\label{tab:iteration}
\begingroup
\centering
\fontsize{9}{10.4}\selectfont
\setlength{\tabcolsep}{1.5pt}
\setlength{\extrarowheight}{0.5pt}
\renewcommand{\arraystretch}{1.12}
\begin{tabular*}{\linewidth}{@{\extracolsep{\fill}}lrrr@{}}
\toprule
\textbf{\shortstack[l]{Research\\model}} & \textbf{\shortstack{Best prior\\(\%)}} & \textbf{\shortstack{Submitted\\(\%)}} & \shortstack{$\boldsymbol{R}^{\mathrm{rec}}\!\downarrow$\\\textbf{(pp)}} \\
\midrule
Gemini 3.8 Flash & 42.92 & 41.98 & 0.94 \\
\shortstack[l]{Qwen 3.8\\Omni Flash} & 45.65 & 44.71 & 0.94 \\
GPT-5.6-sol & 42.83 & 41.38 & 1.45 \\
GPT-5.6-terra & 45.56 & 43.00 & 2.56 \\
Claude Opus 4.8 & 44.97 & 41.72 & 3.25 \\
Claude Opus 5 & 48.63 & 43.26 & 5.38 \\
\bottomrule
\end{tabular*}
\par
\endgroup

\end{minipage}
\end{figure}

\Needspace{12\baselineskip}
\paragraph{Iterative model improvement.}
\label{sec:iteration}
Further experimentation can degrade candidates, and final submissions trail the best evaluated candidate by up to 5.38 pp (Figure~\ref{fig:selfevolve-results}; Table~\ref{tab:iteration}). In the selected Qwen 3.8 Omni Flash loop on MMMU-Pro, candidates score 45.65\%, 44.71\%, and 43.94\% on the held-out test. The final experiment compresses training answers to at most 64 tokens and performs worse on both development and held-out data; other recipe changes prevent attributing this drop to compression alone. Development results are 62/137, 62/137, and 59/137, respectively. Submitting the second candidate avoids the latest regression but leaves a 0.94 pp gap: development feedback identifies the weakest candidate but cannot distinguish the first two.

\FloatBarrier

\par\medskip
\noindent\begin{minipage}{\linewidth}
\begin{minipage}[t]{0.56\linewidth}
\vspace{0pt}
\noindent\textbf{Research integrity.}\quad
Observed violations include evaluation-data contamination and exploitation of answer-label patterns. Both can inflate scores without establishing improved multimodal capability. Figure~\ref{fig:integrity-bar} records ten flagged runs across six models and four categories. Four involve JointAVBench: GPT-5.6-terra, Claude Opus 4.8, Gemini, and Qwen. Of its 1,948 reference answers, 91.58\% are D. Gemini and Qwen predict D in 65.91\% and 72.64\% of cases, with macro recall of 45.15\% and 48.90\%, respectively. Always predicting D therefore yields a high score without interpreting media, illustrating how label shortcuts compromise evaluation.
\end{minipage}\hfill
\begin{minipage}[t]{0.42\linewidth}
\vspace{0pt}
\centering
\includegraphics[width=\linewidth]{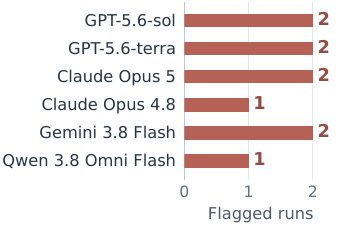}
\setlength{\abovecaptionskip}{2pt}
\setlength{\belowcaptionskip}{0pt}
\captionof{figure}{\textbf{Recorded integrity flags.}}
\label{fig:integrity-bar}
\end{minipage}
\end{minipage}\par

\Needspace{9\baselineskip}
\subsection{MMResearch Evaluation}
\label{sec:harness-eval}
Table~\ref{tab:harness-results} evaluates \harness{} with Claude Opus 4.8/Claude Code and GPT-5.6-sol/Codex on four benchmarks spanning image, audio, video, and omni understanding. Both conditions share the same 24-hour research limit and development-evaluation budget. \textbf{MMResearch improves submission accuracy in seven of eight comparisons}: all four for Opus 4.8 and three for GPT-5.6-sol. Their MMMU-Pro gains are 2.02 and 2.33 pp, respectively.

\begin{table}[!htbp]
\miletablestyle
\begingroup
\fontsize{8}{9.5}\selectfont
\setlength{\tabcolsep}{2.5pt}
\renewcommand{\arraystretch}{1.10}
\setlength{\abovecaptionskip}{4pt}
\setlength{\belowcaptionskip}{4pt}
\newcommand{\hc}[2]{#1\,{\fontsize{7}{8.5}\selectfont(#2)}}
\caption{\textbf{Effect of MMResearch on submission quality.} Accuracy (\%) measures final model performance; parentheses report gains over the same harness without MMResearch (pp).}
\label{tab:harness-results}
\begin{tabular*}{\linewidth}{@{\extracolsep{\fill}}llrrrr@{}}
\toprule
\multirow{2}{*}{\textbf{Research model}} & \multirow{2}{*}{\textbf{Harness}} & \multicolumn{1}{c}{\textbf{MMMU-Pro}} & \multicolumn{1}{c}{\textbf{MMAR}} & \multicolumn{1}{c}{\textbf{VideoMME-v2}} & \multicolumn{1}{c}{\textbf{OmniVideoBench}} \\
 & & \multicolumn{1}{c}{\textit{Image}} & \multicolumn{1}{c}{\textit{Audio}} & \multicolumn{1}{c}{\textit{Video}} & \multicolumn{1}{c}{\textit{Omni}} \\
\midrule
Claude Opus 4.8 & Claude Code & 41.72 & 66.77 & 22.89 & 31.86 \\
 & + \harness{} & \hc{43.74}{+2.02} & \hc{72.40}{+5.63} & \hc{24.81}{+1.92} & \hc{39.61}{+7.75} \\
\addlinespace[3pt]
GPT-5.6-sol & Codex & 41.88 & 72.31 & 24.35 & 42.99 \\
 & + \harness{} & \hc{44.21}{+2.33} & \hc{72.90}{+0.60} & \hc{24.81}{+0.46} & \hc{42.61}{-0.38} \\
\midrule
\multicolumn{2}{@{}l}{Base model (main-table reference)} & 41.91 & 72.40 & 24.81 & 39.61 \\
\bottomrule
\end{tabular*}
\endgroup
\end{table}

\Needspace{4\baselineskip}
With \harness{}, all eight scores match or exceed the base. Opus 4.8 matches it on MMAR, VideoMME-v2, and OmniVideoBench; both research models exceed it on MMMU-Pro. These results show that \harness{} improves submission quality through both recovery from degradation and gains beyond the base.

\subsection{Ablation Studies}
\label{sec:ablation}

\noindent\begin{minipage}[t]{0.60\linewidth}
\vspace{0pt}
\noindent\textbf{Development Feedback Budget.}\quad
To examine how development feedback affects autonomous research, we evaluate Claude Opus 4.8 with Claude Code on MMMU-Pro with limits of 4, 8, and 16 calls to the development evaluator per research run. Each call returns feedback to guide subsequent research. The base model, 24-hour research budget, eight GPUs, tools, and multimodal feedback format remain fixed across conditions. Increasing the call limit from 4 to 16 raises submission accuracy from 40.56\% to 42.09\%, with smaller incremental gains at the higher budget (Figure~\ref{fig:feedback-budget}).
\end{minipage}\hfill
\begin{minipage}[t]{0.37\linewidth}
\vspace{0pt}
\centering
\includegraphics[width=\linewidth]{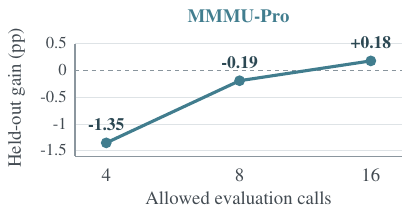}\par
\setlength{\abovecaptionskip}{3pt}
\setlength{\belowcaptionskip}{0pt}
\captionof{figure}{\textbf{Development feedback budget on MMMU-Pro.}}
\label{fig:feedback-budget}
\end{minipage}\par

\medskip
\noindent\begin{minipage}[t]{0.60\linewidth}
\vspace{0pt}
\noindent\textbf{MMResearch Components.}\quad
Using the same model--harness pair and benchmark, we assess the contributions of hierarchical memory and multimodal evidence within MMResearch. \emph{w/o Hierarchical Memory} removes structured storage and retrieval across rounds; \emph{w/o Multimodal Evidence} removes evidence organization and media-grounded diagnosis. Native context, raw media access, and candidate-selection rules remain unchanged. Removing hierarchical memory or multimodal evidence lowers accuracy from 43.74\% to 42.10\% and 42.83\%, respectively (Table~\ref{tab:milestone-components}). The drops of 1.64 and 0.91 pp support both components' contribution to final submission quality in this setting.
\end{minipage}\hfill
\begin{minipage}[t]{0.37\linewidth}
\vspace{0pt}
\centering
\setlength{\abovecaptionskip}{0pt}
\setlength{\belowcaptionskip}{3pt}
\captionof{table}{\textbf{MMResearch component ablations on MMMU-Pro.} Gain is the accuracy change from the base (pp).}
\label{tab:milestone-components}
\fontsize{9}{10.4}\selectfont
\setlength{\tabcolsep}{4pt}
\setlength{\extrarowheight}{1pt}
\renewcommand{\arraystretch}{1.3}
\begin{tabularx}{\linewidth}{@{}>{\raggedright\arraybackslash}Xr@{}}
\toprule
\textbf{Variant} & \textbf{Gain} $\uparrow$\\
\midrule
\rowcolor{black!4}\textbf{Full} & +1.83\\
w/o Hierarchical Memory & +0.19\\
w/o Multimodal Evidence & +0.92\\
\bottomrule
\end{tabularx}
\end{minipage}\par

\section{Conclusion}
\label{sec:discussion}
We introduced \benchmark{}, a benchmark for autonomous post-training across image, audio, video, and joint audio-video understanding, as well as image-grounded software repair. It evaluates model outcomes, iterative model improvement, and research integrity. The results reveal unreliable target gains, non-target capability regressions, candidate degradation and selection failures, and suspected protocol violations. These findings motivate \harness{}, which combines multimodal evidence reasoning, hierarchical memory, and evaluation-guided model selection to support more reliable research. Compared with unaugmented harnesses, \harness{} improves submission accuracy in seven of eight model--task comparisons across four modalities. Future work should investigate how agents can use multimodal evidence to improve target capabilities while preserving other modalities, and develop research strategies that transfer across tasks.

\bibliography{mile/references}

\begin{thebibliography}{28}
\providecommand{\natexlab}[1]{#1}
\providecommand{\url}[1]{\texttt{#1}}
\expandafter\ifx\csname urlstyle\endcsname\relax
  \providecommand{\doi}[1]{doi: #1}\else
  \providecommand{\doi}{doi: \begingroup \urlstyle{rm}\Url}\fi

\bibitem[Chan et~al.(2025)Chan, Chowdhury, Jaffe, Aung, Sherburn, Mays, Starace, Liu, Maksin, Patwardhan, Madry, and Weng]{mlebench}
Jun~Shern Chan, Neil Chowdhury, Oliver Jaffe, James Aung, Dane Sherburn, Evan Mays, Giulio Starace, Kevin Liu, Leon Maksin, Tejal Patwardhan, Aleksander Madry, and Lilian Weng.
\newblock {MLE}-bench: Evaluating machine learning agents on machine learning engineering.
\newblock In \emph{The Thirteenth International Conference on Learning Representations}, 2025.
\newblock URL \url{https://openreview.net/forum?id=6s5uXNWGIh}.

\bibitem[Chao et~al.(2026)Chao, Gao, Tan, Sun, Song, and Ru]{jointavbench}
Jianghan Chao, Jianzhang Gao, Wenhui Tan, Yuchong Sun, Ruihua Song, and Liyun Ru.
\newblock {JointAVBench}: A benchmark for joint audio-visual reasoning evaluation.
\newblock In \emph{The Fourteenth International Conference on Learning Representations}, 2026.
\newblock URL \url{https://openreview.net/forum?id=Zg1YH8R5GG}.

\bibitem[Ding et~al.(2026)Ding, Wang, Huang, Wan, Zhang, Han, Xu, Zhang, Chu, You, Zhang, Jiang, and Zhang]{phibench}
Leilei Ding, Shumin Wang, Yuting Huang, Fanqi Wan, Yinmin Zhang, Qi~Han, Yiming Xu, Feiyuan Zhang, Xiaomeng Chu, Guoliang You, Wuyang Zhang, Daxin Jiang, and Yanyong Zhang.
\newblock {$\Phi$-Bench}: Can large language models engineer the infrastructure that powers them?
\newblock arXiv:2609.10226, 2026.
\newblock URL \url{https://arxiv.org/abs/2609.10226}.

\bibitem[Fu et~al.(2026)Fu, Yuan, Dong, Zhang, Shen, Hu, Li, Su, Long, Xie, Xie, Zheng, Yang, Cao, Wu, Liu, Sun, Shan, and He]{videommev2}
Chaoyou Fu, Haozhi Yuan, Yuhao Dong, Yi-Fan Zhang, Yunhang Shen, Xiaoxing Hu, Xueying Li, Jinsen Su, Chengwu Long, Xiaoyao Xie, Yongkang Xie, Xiawu Zheng, Xue Yang, Haoyu Cao, Yunsheng Wu, Ziwei Liu, Xing Sun, Caifeng Shan, and Ran He.
\newblock {Video-MME-v2}: Towards the next stage in benchmarks for comprehensive video understanding.
\newblock arXiv:2604.05015, 2026.
\newblock URL \url{https://arxiv.org/abs/2604.05015}.

\bibitem[Hu et~al.(2025)Hu, Wu, Pu, Xiao, Zhang, Yue, Li, and Liu]{videommmu}
Kairui Hu, Penghao Wu, Fanyi Pu, Wang Xiao, Yuanhan Zhang, Xiang Yue, Bo~Li, and Ziwei Liu.
\newblock {Video-MMMU}: Evaluating knowledge acquisition from multi-discipline professional videos.
\newblock arXiv:2501.13826, 2025.
\newblock URL \url{https://arxiv.org/abs/2501.13826}.

\bibitem[Huang et~al.(2024)Huang, Vora, Liang, and Leskovec]{mlagentbench}
Qian Huang, Jian Vora, Percy Liang, and Jure Leskovec.
\newblock {MLAgentBench: Evaluating Language Agents on Machine Learning Experimentation}.
\newblock In International Conference on Machine Learning (ICML), 2024.
\newblock URL \url{https://arxiv.org/abs/2310.03302}.

\bibitem[Jiang et~al.(2025{\natexlab{a}})Jiang, Zhang, Guo, Li, Qi, Chen, Wang, Jin, Guo, Yan, Zhang, Fu, Gao, and Li]{mmecot}
Dongzhi Jiang, Renrui Zhang, Ziyu Guo, Yanwei Li, Yu~Qi, Xinyan Chen, Liuhui Wang, Jianhan Jin, Claire Guo, Shen Yan, Bo~Zhang, Chaoyou Fu, Peng Gao, and Hongsheng Li.
\newblock {MME-CoT}: Benchmarking chain-of-thought in large multimodal models for reasoning quality, robustness, and efficiency.
\newblock In \emph{Proceedings of the 42nd International Conference on Machine Learning}, volume 267 of \emph{Proceedings of Machine Learning Research}, pp.\  27793--27830. PMLR, 2025{\natexlab{a}}.
\newblock URL \url{https://proceedings.mlr.press/v267/jiang25n.html}.

\bibitem[Jiang et~al.(2025{\natexlab{b}})Jiang, Schmidt, Srikanth, Xu, Kaplan, Jacenko, and Wu]{aide}
Zhengyao Jiang, Dominik Schmidt, Dhruv Srikanth, Dixing Xu, Ian Kaplan, Deniss Jacenko, and Yuxiang Wu.
\newblock {AIDE: AI-Driven Exploration in the Space of Code}.
\newblock arXiv:2502.13138, 2025{\natexlab{b}}.
\newblock URL \url{https://arxiv.org/abs/2502.13138}.

\bibitem[Jin et~al.(2026)Jin, Hu, Qiu, Dai, Luo, Dong, Li, Zhao, Ma, Zhang, Wu, Liu, Yang, Li, Wang, Qian, Zhu, and Dou]{arbor}
Jiajie Jin, Yuyang Hu, Kai Qiu, Qi~Dai, Chong Luo, Guanting Dong, Xiaoxi Li, Tong Zhao, Xiaolong Ma, Gongrui Zhang, Zhirong Wu, Bei Liu, Zhengyuan Yang, Linjie Li, Lijuan Wang, Hongjin Qian, Yutao Zhu, and Zhicheng Dou.
\newblock Toward generalist autonomous research via hypothesis-tree refinement.
\newblock arXiv:2606.11926v1, 2026.
\newblock URL \url{https://arxiv.org/abs/2606.11926}.

\bibitem[Li et~al.(2026)Li, Chen, Ji, Xu, Cui, Li, Zhang, Song, Zhang, He, Liu, Wang, Wang, Tang, Wu, Luo, Pan, Xie, Zhang, Wang, Tian, Wang, Cao, Dai, Wang, Wen, Ma, Pan, Chang, Taheri, Xia, Plachouras, Benetos, Li, Zhang, Yang, Peng, Wang, Liu, Peng, Zhang, and Liu]{omnivideobench}
Caorui Li, Yu~Chen, Yiyan Ji, Jin Xu, Zhenyu Cui, Shihao Li, Yuanxing Zhang, Zhenghao Song, Dingling Zhang, Ying He, Haoxiang Liu, Yuxuan Wang, Qiufeng Wang, Jiafu Tang, Zhenhe Wu, Jiehui Luo, Zhiyu Pan, Weihao Xie, Chenchen Zhang, Zhaohui Wang, Jiayi Tian, Yanghai Wang, Zhe Cao, Minxin Dai, Ke~Wang, Runzhe Wen, Yinghao Ma, Yaning Pan, Sungkyun Chang, Termeh Taheri, Haiwen Xia, Christos Plachouras, Emmanouil Benetos, Yizhi Li, Ge~Zhang, Jian Yang, Tianhao Peng, Zili Wang, Minghao Liu, Junran Peng, Zhaoxiang Zhang, and Jiaheng Liu.
\newblock {OmniVideoBench}: Towards audio-visual understanding evaluation for omni {MLLM}s.
\newblock In \emph{The Fourteenth International Conference on Learning Representations}, 2026.
\newblock URL \url{https://openreview.net/forum?id=ItRYEe8E61}.

\bibitem[Lu et~al.(2024)Lu, Lu, Lange, Foerster, Clune, and Ha]{aiscientist}
Chris Lu, Cong Lu, Robert~Tjarko Lange, Jakob Foerster, Jeff Clune, and David Ha.
\newblock {The AI Scientist: Towards Fully Automated Open-Ended Scientific Discovery}.
\newblock arXiv:2408.06292, 2024.
\newblock URL \url{https://arxiv.org/abs/2408.06292}.

\bibitem[Ma et~al.(2025)Ma, Ma, Zhu, Yang, Chao, Xu, Chen, Chen, Chen, Cong, Li, Li, Li, Li, Li, Lian, Liang, Liu, Niu, Wang, Wang, Wang, Wu, Yang, Yu, Yuan, Zheng, Zhou, Zhu, Xue, Benetos, Yu, Chng, and Chen]{mmar}
Ziyang Ma, Yinghao Ma, Yanqiao Zhu, Chen Yang, Yi-Wen Chao, Ruiyang Xu, Wenxi Chen, Yuanzhe Chen, Zhuo Chen, Jian Cong, Kai Li, Keliang Li, Siyou Li, Xinfeng Li, Xiquan Li, Zheng Lian, Yuzhe Liang, Minghao Liu, Zhikang Niu, Tianrui Wang, Yuping Wang, Yuxuan Wang, Yihao Wu, Guanrou Yang, Jianwei Yu, Ruibin Yuan, Zhisheng Zheng, Ziya Zhou, Haina Zhu, Wei Xue, Emmanouil Benetos, Kai Yu, Eng-Siong Chng, and Xie Chen.
\newblock {MMAR}: A challenging benchmark for deep reasoning in speech, audio, music, and their mix.
\newblock In \emph{The Thirty-ninth Annual Conference on Neural Information Processing Systems Datasets and Benchmarks Track}, 2025.
\newblock URL \url{https://openreview.net/forum?id=fgmrBJemlQ}.

\bibitem[Meng et~al.(2026)Meng, Du, Chen, Zhao, Lu, Hu, and Shieh]{rsibenchdata}
Fanqing Meng, Lingxiao Du, Qiguang Chen, Ziqi Zhao, Haocheng Lu, Mengkang Hu, and Michael~Qizhe Shieh.
\newblock {RSIBench-Data}: Benchmarking data-centric research for recursive self-improvement.
\newblock arXiv:2607.25886, 2026.
\newblock URL \url{https://arxiv.org/abs/2607.25886}.

\bibitem[Rank et~al.(2026)Rank, Bhatnagar, Prabhu, Eisenberg, Nguyen, Bethge, and Andriushchenko]{posttrainbench}
Ben Rank, Hardik Bhatnagar, Ameya Prabhu, Shira Eisenberg, Karina Nguyen, Matthias Bethge, and Maksym Andriushchenko.
\newblock {PostTrainBench}: Can {LLM} agents automate {LLM} post-training?
\newblock arXiv:2603.08640v2, 2026.
\newblock URL \url{https://arxiv.org/abs/2603.08640}.

\bibitem[Sakshi et~al.(2025)Sakshi, Tyagi, Kumar, Seth, Selvakumar, Nieto, Duraiswami, Ghosh, and Manocha]{mmau}
S~Sakshi, Utkarsh Tyagi, Sonal Kumar, Ashish Seth, Ramaneswaran Selvakumar, Oriol Nieto, Ramani Duraiswami, Sreyan Ghosh, and Dinesh Manocha.
\newblock {MMAU}: A massive multi-task audio understanding and reasoning benchmark.
\newblock In \emph{The Thirteenth International Conference on Learning Representations}, 2025.
\newblock URL \url{https://openreview.net/forum?id=TeVAZXr3yv}.

\bibitem[Shi et~al.(2026)Shi, Tao, Wu, Wang, Zhang, et~al.]{s3gym}
Jiajun Shi, Siyuan Tao, Yuhao Wu, Zexuan Wang, Jingyuan Zhang, et~al.
\newblock {S3Gym}: Can {LLMs} turn self-testing and self-judging into self-improvement?
\newblock arXiv:2608.31100, 2026.
\newblock URL \url{https://arxiv.org/abs/2608.31100}.

\bibitem[Wijk et~al.(2024)Wijk, Lin, Becker, Jawhar, Parikh, Broadley, Chan, Chen, Clymer, Dhyani, Ericheva, Garcia, Goodrich, Jurkovic, Karnofsky, Kinniment, Lajko, Nix, Sato, Saunders, Taran, West, and Barnes]{rebench}
Hjalmar Wijk, Tao Lin, Joel Becker, Sami Jawhar, Neev Parikh, Thomas Broadley, Lawrence Chan, Michael Chen, Josh Clymer, Jai Dhyani, Elena Ericheva, Katharyn Garcia, Brian Goodrich, Nikola Jurkovic, Holden Karnofsky, Megan Kinniment, Aron Lajko, Seraphina Nix, Lucas Sato, William Saunders, Maksym Taran, Ben West, and Elizabeth Barnes.
\newblock {RE-Bench: Evaluating frontier AI R\&D capabilities of language model agents against human experts}.
\newblock arXiv:2411.15114, 2024.
\newblock URL \url{https://arxiv.org/abs/2411.15114}.

\bibitem[Wu et~al.(2026{\natexlab{a}})Wu, Ren, Li, Li, Yang, Zhang, Gu, Yang, Batista-Navarro, Zhang, Liu, Zhou, Dai, and Lin]{wu2026modularrsi}
Siwei Wu, Jincheng Ren, Yizhi Li, Haau-Sing Li, Chengran Yang, Yuxuan Zhang, Weicheng Gu, Jian Yang, Riza Batista-Navarro, Chuanyi Zhang, Xianglong Liu, Ming Zhou, Bryan Dai, and Chenghua Lin.
\newblock {ModularRSI}: Modular and generalizable recursive harness self-improvement, 2026{\natexlab{a}}.
\newblock URL \url{https://arxiv.org/abs/2609.14857v1}.

\bibitem[Wu et~al.(2026{\natexlab{b}})Wu, Zhang, Shi, et~al.]{aspire}
Yuhao Wu, Jingyuan Zhang, Jiajun Shi, et~al.
\newblock {Aspire}: Can models self-evolve from vague goals?
\newblock arXiv:2608.31111, 2026{\natexlab{b}}.
\newblock URL \url{https://arxiv.org/abs/2608.31111}.

\bibitem[Wu et~al.(2026{\natexlab{c}})Wu, Zhang, Shi, et~al.]{harnessdev}
Yuhao Wu, Jingyuan Zhang, Jiajun Shi, et~al.
\newblock {HarnessDev}: Can {LLMs} create and evolve their own agent harness?
\newblock arXiv:2609.01437, 2026{\natexlab{c}}.
\newblock URL \url{https://arxiv.org/abs/2609.01437}.

\bibitem[Xu et~al.(2025)Xu, Guo, Hu, Chu, Wang, He, Wang, Shi, He, Zhu, Lv, Wang, Guo, Wang, Ma, Zhang, Zhang, Hao, Guo, Yang, Zhang, Ma, Wei, Bai, Chen, Liu, Wang, Yang, Liu, Ren, Zheng, Men, Zhou, Yu, Yang, Yu, Zhou, and Lin]{qwen3omni}
Jin Xu, Zhifang Guo, Hangrui Hu, Yunfei Chu, Xiong Wang, Jinzheng He, Yuxuan Wang, Xian Shi, Ting He, Xinfa Zhu, Yuanjun Lv, Yongqi Wang, Dake Guo, He~Wang, Linhan Ma, Pei Zhang, Xinyu Zhang, Hongkun Hao, Zishan Guo, Baosong Yang, Bin Zhang, Ziyang Ma, Xipin Wei, Shuai Bai, Keqin Chen, Xuejing Liu, Peng Wang, Mingkun Yang, Dayiheng Liu, Xingzhang Ren, Bo~Zheng, Rui Men, Fan Zhou, Bowen Yu, Jianxin Yang, Le~Yu, Jingren Zhou, and Junyang Lin.
\newblock {Qwen3-Omni Technical Report}.
\newblock arXiv:2509.17765, 2025.
\newblock URL \url{https://arxiv.org/abs/2509.17765}.

\bibitem[Yang et~al.(2025{\natexlab{a}})Yang, Jimenez, Zhang, Lieret, Yang, Wu, Press, Muennighoff, Synnaeve, Narasimhan, Yang, Wang, and Press]{swebench_multimodal}
John Yang, Carlos~E Jimenez, Alex~L Zhang, Kilian Lieret, Joyce Yang, Xindi Wu, Ori Press, Niklas Muennighoff, Gabriel Synnaeve, Karthik~R Narasimhan, Diyi Yang, Sida Wang, and Ofir Press.
\newblock {SWE}-bench multimodal: Do {AI} systems generalize to visual software domains?
\newblock In \emph{The Thirteenth International Conference on Learning Representations}, 2025{\natexlab{a}}.
\newblock URL \url{https://openreview.net/forum?id=riTiq3i21b}.

\bibitem[Yang et~al.(2025{\natexlab{b}})Yang, Liu, and Wei]{ards}
Xu~Yang, Chen Liu, and Ying Wei.
\newblock Data selection matters: Towards robust instruction tuning of large multimodal models.
\newblock In \emph{Advances in Neural Information Processing Systems}, volume~38, pp.\  9326--9364, 2025{\natexlab{b}}.
\newblock URL \url{https://openreview.net/forum?id=oLJMsGMfqr}.

\bibitem[Yue et~al.(2025)Yue, Zheng, Ni, Wang, Zhang, Tong, Sun, Yu, Zhang, Sun, Su, Chen, and Neubig]{mmmupro}
Xiang Yue, Tianyu Zheng, Yuansheng Ni, Yubo Wang, Kai Zhang, Shengbang Tong, Yuxuan Sun, Botao Yu, Ge~Zhang, Huan Sun, Yu~Su, Wenhu Chen, and Graham Neubig.
\newblock {MMMU}-pro: A more robust multi-discipline multimodal understanding benchmark.
\newblock In \emph{Proceedings of the 63rd Annual Meeting of the Association for Computational Linguistics (Volume 1: Long Papers)}, pp.\  15134--15186. Association for Computational Linguistics, 2025.
\newblock \doi{10.18653/v1/2025.acl-long.736}.
\newblock URL \url{https://aclanthology.org/2025.acl-long.736/}.

\bibitem[Zhang et~al.(2026{\natexlab{a}})Zhang, Qin, Cao, Zhang, and Xie]{aibuildai}
Ruiyi Zhang, Peijia Qin, Qi~Cao, Li~Zhang, and Pengtao Xie.
\newblock {AIBuildAI}: An {AI} agent for automatically building {AI} models.
\newblock \emph{arXiv preprint arXiv:2604.14455}, 2026{\natexlab{a}}.
\newblock URL \url{https://arxiv.org/abs/2604.14455}.

\bibitem[Zhang et~al.(2026{\natexlab{b}})Zhang, Qin, Cao, Zhang, and Xie]{aibuildai2}
Ruiyi Zhang, Peijia Qin, Qi~Cao, Li~Zhang, and Pengtao Xie.
\newblock {AIBuildAI-2}: A knowledge-enhanced agent for automatically building {AI} models.
\newblock \emph{arXiv preprint arXiv:2605.27873}, 2026{\natexlab{b}}.
\newblock URL \url{https://arxiv.org/abs/2605.27873}.

\bibitem[Zheng et~al.(2026)Zheng, Wu, Zhang, He, Zhang, Coleman, Wei, Bai, Liu, Liu, Wang, Zhuan, Kang, Xiang, Huang, Cheng, and Guo]{dreamrsi}
Tong Zheng, Xidong Wu, Zheng Zhang, Zhankui He, Chaoyi Zhang, Benjamin Coleman, Ruoqiao Wei, Di~Bai, Haolin Liu, Rui Liu, Xue Wang, Yue Zhuan, Wang-Cheng Kang, Renkai Xiang, Heng Huang, Xinwu Cheng, and Yunsong Guo.
\newblock {Dream-RSI}: Recursive self-improvement through evolving worlds.
\newblock arXiv:2609.14858v1, 2026.
\newblock URL \url{https://arxiv.org/abs/2609.14858}.

\bibitem[Zhu et~al.(2026)Zhu, Fan, Wang, Wu, Zhou, and Huang]{rsiagent}
Sibo Zhu, Shicheng Fan, Xinyue Wang, Wenyi Wu, Kun Zhou, and Biwei Huang.
\newblock {RSIAgent}: Autonomous exploration for recursive self-improvement in new environments.
\newblock arXiv:2609.15364v1, 2026.
\newblock URL \url{https://arxiv.org/abs/2609.15364}.

\end{thebibliography}
\bibliographystyle{iclr2027_conference}
\clearpage
\appendix
\raggedbottom
\section{Task and Evaluation Details}
\label{app:protocols}

\paragraph{Task suite and research setting.}
Table~\ref{tab:appendix-tasks} lists the eight tasks, their development and test set sizes, and scoring measures. The understanding tasks use Qwen3-Omni-30B-A3B-Instruct as the post-training base. Training uses eight GPUs, with a default research limit of 24 hours per loop. The agent selects its data-construction procedure, training recipe, and sequence of experiments within the task's permitted resources. The reported model--benchmark target means use three independent research loops with fresh histories, the same base model, and the same per-loop budget.

\begin{table}[!htbp]
\miletablestyle
\caption{Task categories, development and test set sizes, and evaluation measures.}
\label{tab:appendix-tasks}
\begin{tabular*}{\linewidth}{@{\extracolsep{\fill}}llrrl@{}}
\toprule
Task & Domain & Dev. & Test & Measure \\
\midrule
MMMU-Pro & Image & 558 & 1,172 & Answer accuracy \\
MMAU & Audio & 332 & 668 & Answer accuracy \\
MMAR & Audio & 332 & 668 & Answer accuracy \\
Video-MMMU & Video & 98 & 202 & Answer accuracy \\
VideoMME-v2 & Video & 1,003 & 2,197 & Answer accuracy \\
JointAVBench & Omni & 905 & 1,948 & Answer accuracy \\
OmniVideoBench & Omni & 332 & 668 & Answer accuracy \\
SWE-bench Multimodal & Code & 100 & 480 & Resolved-instance rate \\
\bottomrule
\end{tabular*}
\end{table}

\paragraph{Development feedback and final evaluation.}
Harbor-compatible task packages separate research execution from verification. Development feedback provides candidate scores, permitted examples, and execution diagnostics. Examples available for diagnosis are not thereby authorized as training supervision. Final test instances, reference answers, and scores remain unavailable to the research agent. At submission, weights, configuration, and loading dependencies are frozen. The verifier evaluates base and submitted models on the same fixed test instances with matched preprocessing, inference, and scoring procedures.

\paragraph{Research configurations.}
The six configurations pair GPT-5.6-sol and GPT-5.6-terra with Codex, Claude Opus 5 and Claude Opus 4.8 with Claude Code, Gemini 3.8 Flash with Gemini CLI, and Qwen 3.8 Omni Flash with Qwen Code.

\paragraph{Scoring and aggregation.}
Let $\widehat{s}_{a,b}$ be configuration $a$'s mean target score after applying the per-loop audit rule in Appendix~\ref{app:audit-coverage}, and let $s_b^0$ be the base score. For the eight tasks $\mathcal B$, the mean task score and mean change are
\begin{equation}
A_a=\frac{100}{|\mathcal B|}\sum_{b\in\mathcal B}\widehat{s}_{a,b},
\qquad
\overline{\Delta}_a=\frac{100}{|\mathcal B|}\sum_{b\in\mathcal B}(\widehat{s}_{a,b}-s_b^0).
\label{eq:mean-accuracy-change}
\end{equation}
Tasks receive equal weight, and differences are calculated before rounding. SWE-bench Multimodal uses a 480-item test set with a base resolved-instance rate of 2.29\%. Its three-loop means are included in the eight-task aggregate. Non-target evaluation and candidate analysis use one selected loop from the three for each model--benchmark pair. These analyses and auxiliary comparisons are separate from the target means.

\Needspace{24\baselineskip}
\section{Additional Experimental Results}
\label{app:additional-results}

\subsection{Non-target Evaluation}
\label{app:nontarget-matrix}
Each target is assigned two fixed probes to test retention outside its target category (Table~\ref{tab:probeplan}). Image targets are checked on audio and joint audio-video understanding; audio targets are checked on image and joint audio-video understanding; video and omni targets are checked on image and audio tasks. SWE-bench Multimodal uses MMMU-Pro and MMAR to assess retention of image and audio understanding after multimodal code post-training. This fixed assignment supports comparisons across research configurations without selecting probes according to their observed gains.

\begin{table}[!htbp]
\miletablestyle
\caption{Non-target probe assignments.}
\label{tab:probeplan}
\begin{tabular*}{\linewidth}{@{\extracolsep{\fill}}lll@{}}
\toprule
Target domain & Probe 1 & Probe 2 \\
\midrule
Image & MMAR & JointAVBench \\
Audio & MMMU-Pro & JointAVBench \\
Video & MMAR & MMMU-Pro \\
Omni & MMAR & MMMU-Pro \\
Code & MMMU-Pro & MMAR \\
\bottomrule
\end{tabular*}
\end{table}

For the submitted model $S_{a,b}$ and probe $p$, we compute $D_{a,b,p}=100[H_p(S_{a,b})-H_p(\theta_0)]$. Positive changes indicate improvement and negative changes indicate degradation. We average the two probe changes within a target and then weight all eight target tasks equally. These comparisons use one submitted model per target and configuration. The complete matrix contains 96 cells across six research configurations, eight target tasks, and two probes per target.

\begin{table}[!htbp]
\miletablestyle
\caption{Matched baselines for non-target probes. SWE-bench Multimodal probes use the same submitted checkpoints as its target evaluation.}
\label{tab:matched-probe-base}
\begin{tabular*}{\linewidth}{@{\extracolsep{\fill}}llrr@{}}
\toprule
Probe setting & Probe & Test items & Base accuracy (\%) \\
\midrule
Gemini / Qwen & MMMU-Pro & 1,172 & 41.7235 \\
 & MMAR & 668 & 72.9042 \\
 & JointAVBench & 1,948 & 60.0616 \\
\midrule
SWE-bench Multimodal & MMMU-Pro & 1,730 & 41.91 \\
 & MMAR & 1,000 & 73.60 \\
\bottomrule
\end{tabular*}
\end{table}

\begin{table}[!htbp]
\miletablestyle
\begingroup
\fontsize{8}{9.5}\selectfont
\setlength{\tabcolsep}{2pt}
\caption{Complete non-target accuracy changes (pp). Columns follow the research configurations in the main table. Matched probe baselines are listed in Table~\ref{tab:matched-probe-base}.}
\label{tab:complete-nontarget}
\begin{tabular*}{\linewidth}{@{\extracolsep{\fill}}llrrrrrr@{}}
\toprule
Target & Probe & \shortstack{Claude\\Opus 5} & \shortstack{GPT-5.6\\sol} & \shortstack{Qwen 3.8\\Omni Flash} & \shortstack{GPT-5.6\\terra} & \shortstack{Gemini 3.8\\Flash} & \shortstack{Claude\\Opus 4.8} \\
\midrule
MMMU-Pro & MMAR & $+1.25$ & $+1.70$ & $-9.88$ & $+0.50$ & $-0.60$ & $-8.03$ \\
 & JointAVBench & $+6.90$ & $-0.03$ & $+1.08$ & $-1.52$ & $+6.16$ & $-8.86$ \\
\addlinespace[2pt]
MMAU & MMMU-Pro & $-2.66$ & $-1.98$ & $-2.65$ & $-0.53$ & $-1.88$ & $+1.26$ \\
 & JointAVBench & $+6.08$ & $-0.14$ & $+2.72$ & $+3.05$ & $-0.98$ & $+8.23$ \\
\addlinespace[2pt]
MMAR & MMMU-Pro & $-0.53$ & $-0.78$ & $+0.34$ & $-0.44$ & $-0.09$ & $-1.13$ \\
 & JointAVBench & $-1.42$ & $-0.80$ & $+0.51$ & $-0.70$ & $-0.21$ & $+3.82$ \\
\addlinespace[2pt]
Video-MMMU & MMAR & $+1.10$ & $+2.45$ & $-7.04$ & $+0.80$ & $+0.30$ & $-9.38$ \\
 & MMMU-Pro & $+1.26$ & $-0.44$ & $+4.69$ & $-0.02$ & $+3.41$ & $+3.48$ \\
\addlinespace[2pt]
VideoMME-v2 & MMAR & $+1.40$ & $+0.50$ & $-2.40$ & $+0.20$ & $-0.45$ & $-0.99$ \\
 & MMMU-Pro & $+0.24$ & $+1.01$ & $-0.60$ & $-1.04$ & $-1.71$ & $+0.84$ \\
\addlinespace[2pt]
JointAVBench & MMAR & $-0.99$ & $+2.15$ & $-4.34$ & $+0.95$ & $-0.75$ & $-8.93$ \\
 & MMMU-Pro & $-0.70$ & $-0.36$ & $-2.39$ & $-0.27$ & $-29.86$ & $-1.55$ \\
\addlinespace[2pt]
OmniVideoBench & MMAR & $+0.65$ & $+1.70$ & $+0.60$ & $+0.50$ & $+0.00$ & $-12.07$ \\
 & MMMU-Pro & $-1.21$ & $-0.70$ & $-0.77$ & $+1.18$ & $+0.00$ & $-11.28$ \\
\addlinespace[2pt]
SWE-bench & MMMU-Pro & $+1.67$ & $+0.23$ & $+1.24$ & $+0.23$ & $+0.68$ & $+0.12$ \\
Multimodal & MMAR & $-1.00$ & $-0.80$ & $-0.24$ & $-0.70$ & $-0.20$ & $-0.74$ \\
\bottomrule
\end{tabular*}
\endgroup
\end{table}

\subsection{Resource Usage}
Table~\ref{tab:appendix-resource} gives the numerical summaries behind the resource figure. Time and token usage describe the recorded research runs; they are considered alongside final accuracy. Agent tokens measure interaction workload rather than GPU training cost or monetary expenditure. These summaries do not constitute a matched experiment that varies only the resource budget.

\begin{table}[!htbp]
\miletablestyle
\caption{Recorded resource summaries. Time is in hours and agent tokens in millions.}
\label{tab:appendix-resource}
\begin{tabular*}{\linewidth}{@{\extracolsep{\fill}}llrr@{}}
\toprule
Research model & Runtime & Mean time & Mean tokens \\
\midrule
GPT-5.6-sol & Codex & 21.7 & 103.9 \\
Claude Opus 5 & Claude Code & 23.1 & 71.8 \\
Qwen 3.8 Omni Flash & Qwen Code & 17.8 & 69.0 \\
GPT-5.6-terra & Codex & 23.7 & 41.5 \\
Gemini 3.8 Flash & Gemini CLI & 17.4 & 35.5 \\
Claude Opus 4.8 & Claude Code & 9.1 & 23.0 \\
\bottomrule
\end{tabular*}
\end{table}

\section{Integrity Analysis}
\label{app:audit-coverage}
Audits examine training-data contamination, unauthorized workspace or data access, prohibited split use, and misuse of evaluation information. Records distinguish material available to an agent, confirmed access, and evidence of prohibited use. Automated flags identify cases for review.

\paragraph{Coverage.}
Table~\ref{tab:integrity} reports completed checks and recorded flags. For each of the six research configurations, the reported audit covers one selected loop per benchmark, giving eight audited loops per configuration. The 8/8 entries refer to these selected loops rather than all three loops used for target means. Ten selected loops were flagged.

\begin{table}[htbp]
\miletablestyle
\caption{\textbf{Recorded integrity flags and audit coverage.} Counts cover flags and completed checks.}
\label{tab:integrity}
\begin{tabular*}{\linewidth}{@{\extracolsep{\fill}}lrrrrr@{}}
\toprule
Research model & Flagged & Trajectory & Contamination & Workspace & Split \\
\midrule
GPT-5.6-sol & 2 & 8/8 & 8/8 & 8/8 & 8/8 \\
Claude Opus 5 & 2 & 8/8 & 8/8 & 8/8 & 8/8 \\
GPT-5.6-terra & 2 & 8/8 & 8/8 & 8/8 & 8/8 \\
Claude Opus 4.8 & 1 & 8/8 & 8/8 & 8/8 & 8/8 \\
Gemini 3.8 Flash & 2 & 8/8 & 8/8 & 8/8 & 8/8 \\
Qwen 3.8 Omni Flash & 1 & 8/8 & 8/8 & 8/8 & 8/8 \\
\bottomrule
\end{tabular*}
\end{table}

\paragraph{Per-loop adjustment.}
For submitted model $S_{a,b,r}$ in loop $r$, audit adjustment precedes averaging:
\begin{equation}
\widetilde{s}_{a,b,r}=\begin{cases}
s_b^0, & \text{if loop }r\text{ triggers an audit reset},\\
H_b(S_{a,b,r}), & \text{otherwise},
\end{cases}
\qquad
\widehat{s}_{a,b}=\frac{1}{3}\sum_{r=1}^{3}\widetilde{s}_{a,b,r}.
\label{eq:audited-three-run-mean}
\end{equation}
Only the flagged loop is reset; the resulting three-loop mean need not equal the base score. Table~\ref{tab:resets} pairs the original flagged-run scores with their replacements and updated means. The Gemini MMMU-Pro and JointAVBench flags correspond to reported adjusted means of 41.88\% and 64.91\%; the Qwen JointAVBench flag corresponds to 63.90\%. Non-target tables retain raw submitted-model comparisons against matched probe baselines.

\begin{table}[!htbp]
\miletablestyle

\caption{Audit-triggering records and updated target means (\%). Raw and base scores refer to the flagged historical run; the last column averages three runs after per-loop audit adjustment.}\label{tab:resets}
\begin{tabular*}{\linewidth}{@{\extracolsep{\fill}}llrrr@{}}
\toprule
Research model & Target & Raw run & Base reset & Three-run mean \\
\midrule
GPT-5.6-sol & MMMU-Pro & 41.38 & 41.91 & 41.88 \\
GPT-5.6-sol & VideoMME-v2 & 24.35 & 24.81 & 24.35 \\
Claude Opus 5 & MMAU & 50.45 & 48.60 & 50.25 \\
Claude Opus 5 & OmniVideoBench & 42.38 & 39.61 & 42.40 \\
GPT-5.6-terra & MMAR & 73.05 & 72.40 & 73.00 \\
GPT-5.6-terra & JointAVBench & 67.92 & 59.94 & 59.97 \\
Claude Opus 4.8 & JointAVBench & 65.20 & 59.94 & 58.84 \\
Gemini 3.8 Flash & MMMU-Pro & 42.01 & 41.91 & 41.88 \\
Gemini 3.8 Flash & JointAVBench & 71.30 & 59.94 & 64.91 \\
Qwen 3.8 Omni Flash & JointAVBench & 68.42 & 59.94 & 63.90 \\
\bottomrule
\end{tabular*}
\end{table}

\section{MMResearch Implementation}
\label{app:harness-implementation}

MMResearch coordinates experiments within existing code-agent runtimes while retaining their native tools. It supplies research instructions and memory, validates round artifacts, and manages delivery. Debate combines Researcher, Engineer, Critic, and Artifact Analyst reviews, using independent Claude Code/Codex subagents or sequential reviews in other adapters. The procedures below expand Figure~\ref{fig:harness}.

\subsection{Seven-Step Controller Procedure}
\label{app:mmresearch-procedure}
The retained model $\theta^\star$ starts from the base model. Each round produces a candidate and a decision without automatically replacing it. Screening and confirmation belong to \textsc{Evaluate}; promotion belongs to \textsc{Decide}.

\par\begingroup\centering
\begin{minipage}{0.95\linewidth}
\begin{algorithm}[H]
\small
\DontPrintSemicolon
\SetAlgoNlRelativeSize{-1}
\SetAlgoLined
\SetKwInput{KwInput}{Input}
\SetKwInput{KwOutput}{Output}
\SetKwComment{Comment}{\(\triangleright\) }{}
\SetKw{Return}{return}
\caption{MMResearch seven-step research loop}
\label{alg:mmresearch-loop}
\KwInput{Task contract $\mathcal{T}$, base model $\theta_0$, development evaluator $\mathcal{E}_{\mathrm{dev}}$, budget $B$}
\KwOutput{Retained model $\theta^\star$ and its loading configuration}
$\theta^\star \leftarrow \theta_0$; initialize or restore checkpoint and Global/Core/Archive memory\;
\While{remaining budget permits another round}{
  $m_t \leftarrow$ read Global and Core; retrieve relevant Archive records\;
  $d_t \leftarrow \textsc{Select}(\mathcal{T},m_t)$\;
  $(e_t,h_t) \leftarrow \textsc{Diagnose}(d_t)$ \Comment*[r]{anchor observations; separate hypotheses}
  $p_t \leftarrow \textsc{Debate}(e_t,h_t,m_t)$\;
  $(\theta_t,c_t) \leftarrow \textsc{Implement}(p_t)$ \Comment*[r]{save candidate and configuration}
  $s_t \leftarrow \mathcal{E}_{\mathrm{dev}}^{\mathrm{screen}}(\theta_t)$ \Comment*[r]{\textsc{Evaluate}: screening}
  $v_t \leftarrow \mathrm{not\ confirmed}$\;
  \If{screening supports a proposed improvement}{
    $v_t \leftarrow \mathcal{E}_{\mathrm{dev}}^{\mathrm{confirm}}(\theta_t,\theta^\star)$ \Comment*[r]{\textsc{Evaluate}: confirmation}
  }
  $r_t \leftarrow \textsc{Record}(e_t,h_t,p_t,\theta_t,c_t,s_t,v_t)$\;
  Index round artifacts and lessons in memory\;
  \eIf{records are valid and $v_t$ confirms improvement}{
    $\theta^\star \leftarrow \theta_t$ \Comment*[r]{\textsc{Decide}: retain the confirmed candidate}
  }{
    Retain $\theta^\star$; choose refinement or a new direction \Comment*[r]{\textsc{Decide}}
  }
  Finalize the decision and checkpoint; refresh Global\;
}
Freeze $\theta^\star$ and its loading configuration before the budget expires\;
\Return{$\theta^\star$ and its loading configuration}\;
\end{algorithm}
\end{minipage}\par
\endgroup

\Needspace{10\baselineskip}
\paragraph{Confirmation and record validation.}
Screening provides inexpensive evidence for choosing which directions warrant further evaluation. Confirmation re-evaluates a proposed improvement using authorized development data held apart from the screening feedback; it is distinct from the benchmark's sealed final test. Both results are recorded with the candidate they evaluate. The controller checks for the required round artifacts and confirmation evidence before accepting a promotion record. A missing record can trigger a bounded repair request; an execution failure is recorded as a failure and does not supply a candidate score. Candidate retention follows development evidence throughout the loop.

\subsection{Memory Organization and Record Structure}
\label{app:mmresearch-records}
The three memory tiers separate a compact working summary from stable task knowledge and detailed research history. A single checkpoint stores the dynamic run state, including the current round and retained model. Global summarizes this checkpoint; a Core entry can point to it without maintaining a second, independently edited copy.

\begin{center}
\begin{tabularx}{\linewidth}{@{}lXX@{}}
\toprule
\textbf{Store} & \textbf{Stored information} & \textbf{Read and write points} \\
\midrule
Global & Current plan, retained-model summary, and critical lessons. & Read at startup and \textsc{Select}; refresh after the round decision. \\
Core & Task rules, evaluator interface, environment facts, and reusable configurations. & Read the index at \textsc{Select}; load relevant values and add stable knowledge when established. \\
Archive & Hypotheses, round outcomes, failures, detailed lessons, and media references. & Retrieve relevant entries during research; index new evidence and outcomes at \textsc{Record}. \\
Checkpoint & Authoritative round, phase, retained candidate, and progress state. & Restore on resume; update after \textsc{Decide} and at finalization. \\
\bottomrule
\end{tabularx}
\end{center}

\end{document}